\documentclass[10pt,twocolumn,letterpaper]{article}

\usepackage[pagenumbers]{wacv} 

\usepackage{graphicx}
\usepackage{amsmath}
\usepackage{amssymb}
\usepackage{booktabs}
\usepackage{multirow,makecell}
\usepackage{bm}

\usepackage[pagebackref,breaklinks,colorlinks]{hyperref}

\usepackage[capitalize]{cleveref}
\crefname{section}{Sec.}{Secs.}
\Crefname{section}{Section}{Sections}
\Crefname{table}{Table}{Tables}
\crefname{table}{Tab.}{Tabs.}

\def\wacvPaperID{278} 
\def\confName{WACV}
\def\confYear{2024}

\begin{document}

\title{Can Prompt Learning Benefit Radiology Report Generation?}

\author{Jun Wang$^1$, Lixing Zhu$^{2,3}$, Abhir Bhalerao$^1$ and Yulan He$^{1,2,3}$\\
$^1$University of Warwick, UK; $^2$King's College London, UK \\
$^3$The Alan Turing Institute, UK\\
}

\maketitle

\begin{abstract}
    Radiology report generation aims to automatically provide clinically meaningful descriptions of radiology images such as MRI and X-ray. Although great success has been achieved in natural scene image captioning tasks, radiology report generation remains challenging and requires prior medical knowledge. In this paper, we propose PromptRRG, a method that utilizes prompt learning to activate a pretrained model and incorporate prior knowledge. Since prompt learning for radiology report generation has not been explored before, we begin with investigating prompt designs and categorise them based on varying levels of knowledge: common, domain-specific and disease-enriched prompts. Additionally, we propose an automatic prompt learning mechanism to alleviate the burden of manual prompt engineering. This is the first work to systematically examine the effectiveness of prompt learning for radiology report generation. Experimental results on the largest radiology report generation benchmark, MIMIC-CXR, demonstrate that our proposed method achieves state-of-the-art performance. Code will be available upon the acceptance.
\end{abstract}

\section{Introduction}

\begin{figure}
    \centering     \includegraphics[width=0.45\textwidth]{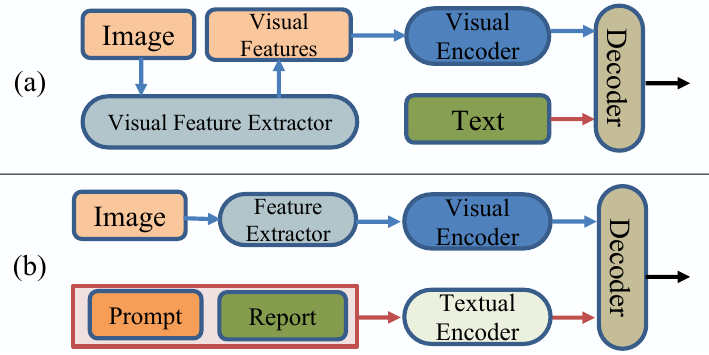} 
    \caption{ (a): The general architectures of recent image captioning/RRG methods. (b): A simplified illustration of the architecture of proposed PromptRRG with manual prompt learning. }    
    \label{fig:intro_motivation}  
    \vspace{-20pt}
\end{figure}

The purpose of radiology report generation (RRG) is to automatically generate clinical descriptions of radiology images such as X-Ray and MRI. Recently, RRG has gained increasing attention due to its huge potential to aid the diagnostic decision process and mitigate the shortage of professional radiologists. Owing to the release of large well-annotated datasets and developments in hardware, a significant improvement has been seen in image captioning methods.  
\autoref{fig:intro_motivation} (a) illustrates this typical architecture, which comprises a pretrained visual extractor that first extracts a grid of features in the image. Subsequently, these feature representations are sent to the encoder, whose outputs, i.e., attended visual tokens, are combined with embedded textual features to serve as the input to the decoder for the autoregressive generation of the captions. 

Notably, recent approaches ~\cite{cornia2020meshed,guo2020normalized,pan2020x,ji2021improving,zeng2022progressive} usually adopt the Transformer~\cite{vaswani2017attention} as the backbone for the encoder-decoder architecture because of its self-attention mechanism and prodigious capability of modelling long-range dependency. Inspired by the success in image captioning, recent state-of-the-art RRG methods~\cite{chen2020generating,chen2021cross,wang2022cross,liu-etal-2021-competence} have adopted a similar general architecture, i.e., encoder-decoder pairs, because of the apparent similarity between these two tasks. However, RRG is far more challenging and a large performance gap can be seen between the widely-used image captioning and RRG benchmarks MIMIC-CXR~\cite{johnson2019mimic} and COCO \cite{lin2014microsoft}, e.g., 11 vs 42 on BLEU4 score, 14 vs 33 on METEOR scores. This is because, image captioning mostly requires generating 1 or 2 sentences, whereas RRG requires 2-4 times more sentences to elaborately describe the findings in an image with more sophisticated semantic relationships and their corresponding image regions. The problem is exacerbated as there exist data biases in commonly used datasets with significantly fewer radiology reports from X-ray images containing abnormal regions. Such abnormal regions are typically only a small proportion of pathological images, making it hard for the RRG models to efficiently capture abnormal features. Furthermore, even in pathological cases, most report sentences may be associated with a description of normal findings. Therefore, generating an accurate radiology report is complicated and needing more informative and semantic knowledge, which is a prerequisite for a model to capture the fine-grained medical information present and the relationships therein.

Previous approaches usually distil prior knowledge into the model by leveraging the information, i.e., reports from similar images, disease tags and medical entities, to form a knowledge graph~\cite{li2019knowledge,zhang2020radiology,kale2023kgvl} or attention-based modules~\cite{liu2021exploring,liu2021contrastive}. Some works~\cite{wang2022cross,chen2021cross} utilize an extra memory matrix to learn and record the cross-modal knowledge. These methods usually embed the reports through an embedding layer without a specific text encoder, and then directly send them into the decoder to perform the cross-modal fusion and report generation, ignoring the power a pretrained language model (PLM) might afford.

Unlike previous works that normally learn the knowledge representation from scratch and require a carefully-designed knowledge learning module with various additional input sources such as reports from other images, we propose to take advantage of existing knowledge in a PLM and employ prompt learning to distil this informative knowledge into a generative model as illustrated in \autoref{fig:intro_motivation} (b) and \autoref{fig:architecture}. Although prompt learning has demonstrated potent capability of fine-tuning pre-trained models for downstream tasks, only a few studies explore it in captioning tasks. In particular, leveraging prompt learning to improve the accuracy of RRG has never been seen in previous studies. Specifically, we begin with the prompt design and propose three different-levels of prompts with increasing amounts of domain knowledge: common prompts, domain-related prompts and disease-enriched prompts. These prompts are taken as the prefix of the report to form the prompt learning. We also present an automatic prompt learning method that enables the model to learn which prompts are useful during the training, without the need for manual prompt engineering.

This work makes three principal contributions:
\begin{enumerate}
    \item A generative, prompt-learning based RRG model, PromtRRG, which leverages prompt learning and pre-trained language models to successfully distil informative knowledge into the RRG task without adding trainable parameters\footnote{For manual prompt learning.} or building auxiliary cross-domain knowledge learning modules.
    \item Proposals for different-levels of domain knowledge prompts and a systematic investigation of their performance.
    \item A method for prompt learning to automatically learn the most effective prompts.
\end{enumerate}
After a detailed description of our proposed methods, we present results of experiments on the MIMIC-CXR benchmark and demonstrate that PromptRRG achieves results that surpass the previous state-of-the-art methods. Discussion and proposals are given to inspire future work.

\section{Related Work}

\subsection{Radiology Report Generation}
Following on from the success in image captioning, recent state-of-the-art RRG studies adopt similar architectures: an encoder-decoder to generate the report in an end-to-end manner.  
For example, Chen et al.~\cite{chen2020generating} adopt this typical architecture and designed a relational memory with a memory-driven conditional layer normalization to better learn the report patterns. R2GenCMN~\cite{chen2021cross} propose a shared memory matrix to implicitly learn the cross-modal alignment. Wang et al.~\cite{wang2022cross} further extend R2GenCMN by integrating the cross-modal prototype learning into the network to improve cross-modal pattern modelling. 

Some studies introduce prior knowledge to help the model better capture medical information. For instance, works that have utilized disease topic information~\cite{liu2019clinically, zhang2020radiology} demonstrate better results. Liu et al.~\cite{liu2021exploring} extended these works by taking advantage of both prior and posterior knowledge, i.e. knowledge from similar images and a predefined topic bag for report generation. However, these works normally learn the knowledge representation from scratch and require a carefully-designed knowledge learning module with various extra input sources such as reports from other images. Instead, we propose to take advantage of the existing knowledge in a PLM and prompt learning to distil the informative knowledge, without introducing extra trainable parameters.

\subsection{Prompt Learning}


Proposed by the GPT series~\cite{brown2020language}, prompt learning has established itself as a powerful technique in natural language processing to navigate agnostic tasks, unifying them as either free-form generation or multiple-choice selection~\cite{LLaMAnotarxivyet}. In a similar vein, computer vision tasks adapt prompt learning to improve pre-trained vision-language models in zero-shot or few-shot settings.
These methods \cite{dou2022empirical,jin2022good,lu2022prompt,bach-etal-2022-promptsource,NEURIPS2022_960a172b,zhou2023large}, covering the vast majority of CV and NLP tasks, predominantly focus on categorization, e.g., image classification, visual-question answering, NLP zero-shot task generalization, and image retrieval.

In contrast, very few works target prompted image-to-text generation. While there are recent ones utilizing prompt learning in image captioning to stylize the generated captions \cite{wang2022controllable} or improve zero-shot predictions \cite{wang2021simvlm,li2022blip},
none of them has explored prompt learning in RRG, a more challenging task that demands coherent long text generation and medical domain induction. To fill in the blanks, we introduce prompted RRG to generate clinically coherent descriptions and precise diagnoses. Unlike methods in image captioning, we concentrate on eliciting the knowledge for higher performance via prompt learning since accuracy is the first priority in the medical domain. To this end, we carefully design some manual prompts with different levels of knowledge.

There are some recent attempts \cite{wang2022controllable,zhou2022learning} on automatic prompt learning to directly optimize the prompt vectors in the continuous space. However, they expect the model can learn the whole embedding procedure such as the word embedding, position embedding, dropout and layer normalization, which is a tricky task and ignores the important position information. We therefore propose a new automatic prompt learning to overcome these problems.

\begin{figure*}
    \centering 
    \includegraphics[width=0.9\textwidth]{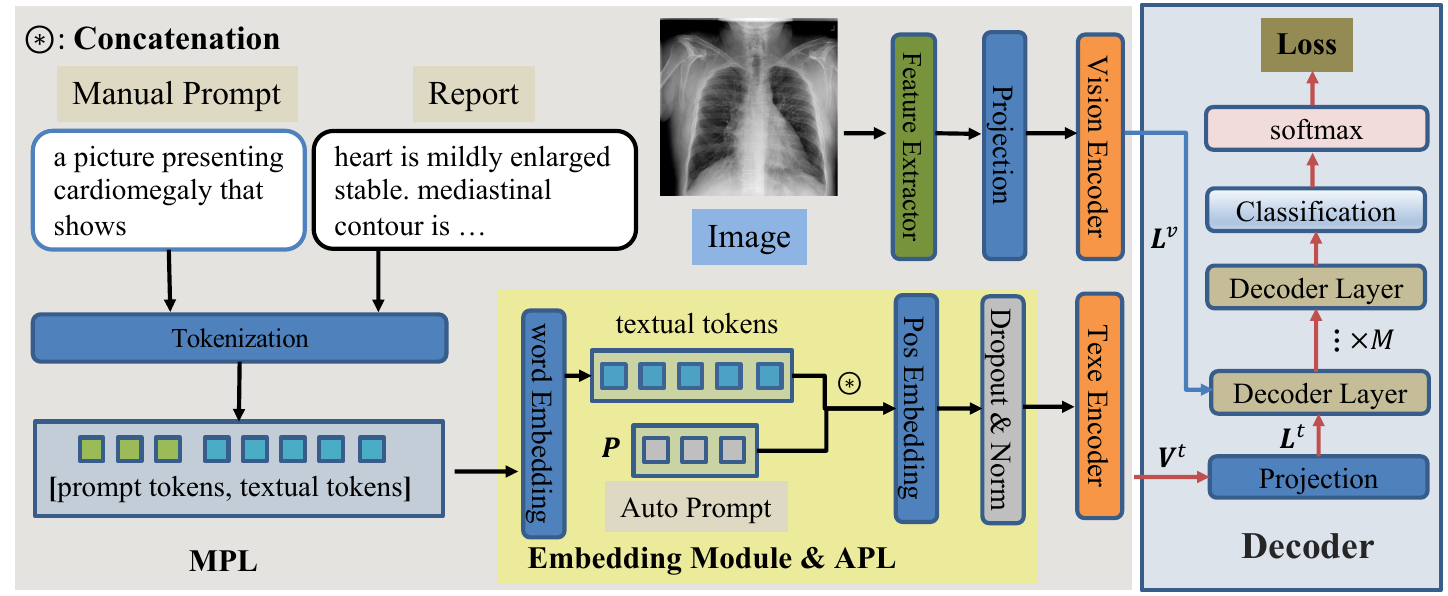} 
    \caption{The illustration of the new framework for RRG with prompt learning. `MPL' and `APL' are the abbreviations of Manual Prompt Learning and Automatic Prompt Learning. }    
    \label{fig:architecture}  
    \vspace{-10pt}
\end{figure*}

\section{Methods}
\label{sec:method}

\subsection{Model Architecture}
Given a radiology image $\bm{I}$, the purpose of RRG is to generate coherent findings (report) $\bm{R}=\{w1,w2,...,w_{N_r}\}$ from $\bm{I}$. Note that $N_r$ is the number of words in the report. The following subsections present the details of the model architecture as illustrated in~\autoref{fig:architecture}.
\subsubsection{Visual Feature Extractor}
In particular, a visual feature extractor, e.g. ResNet101 \cite{he2016deep} or Swin Transformer~\cite{liu2021swin}, is first employed to extract visual features $\bm{V}^i \in \mathbb{R}^{H \times W \times C}$ which are then flattened to a sequence of visual tokens $\bm{V}^s \in \mathbb{R}^{HW \times C}$.  $H$, $W$, $C$ are the height, width and the number of channels respectively. This process is formulated as:
\begin{equation}
\label{eq:vt}
 \{ v^s_1, v^s_2, ...,v^s_k, ..., v^s_{N^s-1}, v^s_{N^s} \} = f_{vfe}(\bm{I}),
\end{equation}
\noindent
where $v^s_k$ denotes the patch feature in the $k^{th}$ position in $\bm{V}^s$, and $N^s=H\times W$. $f_{vfe}$ is the visual feature extractor.

\subsubsection{Encoder}
Different from previous works that learn the word embedding from scratch via an embedding layer, we propose to directly utilize a well-trained semantic embedding by integrating pretrained language models, e.g. Bert \cite{kenton2019bert} or Roberta\cite{liu2019roberta}, into the model as the text encoder. Therefore, our encoder contains a vision encoder trained from scratch and a pretrained text encoder. Both of them consist of several transformer layers~\cite{vaswani2017attention} whose core is the attention mechanism. 

\noindent \textbf{Text Encoder} Specifically, reports are first tokenized to $N^t$ textual tokens $\bm{R}^{\prime}=\{y_1,y_2,...,y_{N^t}\}$ by a byte-pair encoding tokenizer from the pretrained language model where each element in $\bm{R}^{\prime}$ is a unique numeric ID. Each textual token is then embedded by an embedding module 
formulated as:
\begin{align}
\label{eq:word_embed}
 o_i =&\ \bm{W}_{we}^T \cdot y_i, \\
 \label{eq:pos_embed}
 u_i =&\ \bm{W}_{pos}^T \cdot Pos(y_i), \\
 \label{eq:dropln_embed}
 e_i =&\  Dropout(LayerNorm(o_i+u_i)),
\end{align}
where $\bm{W}_{we} \in \mathbb{R}^{N_{we} \times D_t}$ and $\bm{W}_{pos} \in \mathbb{R}^{N_{pos} \times D_t}$ are the embedding matrix for the word embedding and position embedding. $*^T$ refers to the transpose operation. $N_{we}$ and $N_{pos}$ refer to the vocabulary size and the maximum position set in the PLM respectively. $D_t$ denotes the embedding dimension. $Pos(e_i)$  extracts the position index of $e_i$ and reforms it as a one-hot vector. The final embedding of $i^{th}$ textual token $e^t_i$ is obtained by adding the word embedding $o_i$ and position embedding $u_i$ followed by a layer norm and a dropout operations. After that, the textual token embeddings are fed into the text encoder, i.e. a PLM, $f_{te}$ to obtain the textual tokens $\bm{V^t} \in \mathbb{R}^{N_t \times D_t}$ expressed as:
\begin{equation}
\label{eq:te}
 \{ v^t_1, v^t_2, ..., v^t_{N^t} \} = f_{te}(\{e_1,e_2,...,e_{N^t}\}).
\end{equation}
Note that previous studies in RRG normally do not employ the text encoder as shown in \autoref{fig:intro_motivation} (a). Nonetheless, a PLM usually shows better textual encoding capability. Moreover, our prompt learning cannot be applied without a pretrained text encoder. After obtaining the encoded textual features, we then map them into the intermediate latent space via a fully-connected layer as express as~\autoref{eq:map_tf}:
\begin{equation}
\label{eq:map_tf}
\quad l^t_i =\ \bm{W}_{t}^T \cdot v^t_i,
\end{equation}
where $\bm{W}_{t} \in \mathbb{R}^{D_t \times D}$ is the weight matrix and $v^t_i$ denotes $i^{th}$ encoded textual tokens. 

\noindent \textbf{Vision Encoder} Similar to the text encoder, we first map the extracted visual tokens into the same intermediate latent space. To improve the robustness, a batch normalization is adopted followed by a dropout and activation function. This process is summarized as:
\begin{align}
\label{eq:trans_vf}
\{ v^{s*}_1, v^{s*}_2, ..., &v^{s*}_{N^s} \} = BN(\{v^s_1, v^s_2, ..., v^s_{N^s} \}), \\
\label{eq:map_vf}
l^s_i =\ &Dropout(\sigma(\bm{W}_{s}^T \cdot v^{s*}_i)),
\end{align}
where $BN$ and $\sigma$ refer to the batch normalization and ReLU activation respectively. $\bm{W}_{s} \in \mathbb{R}^{D_v \times D}$ is the weight matrix and $v^t_i$ denotes $i^{th}$ visual tokens. These visual tokens are then fed into the vision encoder to perform feature encoding and long-term dependence modelling. 
\begin{equation}
\label{eq:ve}
 \{ l^v_1,l^v_2,...,l^v_{N^s} \} = f_{ve}(\{l^s_1,l^s_2,...,l^s_{N^s}\}).
\end{equation}

\subsubsection{Decoder} 
\label{sec:decoder}
The decoder also consists of several transformer layers. The only difference from the encoder is that a cross-attention is added after the self-attention for the textual tokens to perform the cross-modal feature interaction. Specifically, the decoder takes these two sequences of encoded tokens as source inputs and predicts the current output (word). Representing the decoder as $f_{de}$, the decoding procedure can be then summarized as:
\begin{equation}
\label{eq:decoding}
{p_T}=f_{de}(\{l^v_1,l^v_2,...,l^v_{N_v}\},\{l^t_1,l^t_2,...,l^t_T\}),
\end{equation}
where ${p_T}$ denotes the predicted token in time step $T$. The whole architecture is trained end-to-end by a cross-entropy loss $\mathcal{L}_{CE}$:
\begin{align}
\label{eq:objective_base}
\mathcal{L}_{CE}=-\sum_T \log P(y_T |f_{de}(\bm{L}^v,\bm{L}^{t}_{<T})),
\end{align}
where $\bm{L}^v$ and $\bm{L}^t$ are the visual and textual tokens from the encoder.

\subsection{RRG with Prompt Learning}
\label{sec:prompt_learning}
It has been proved that a good prompt has the ability to trigger the memories of the pre-trained language model. In this work, we expect that the prompt can convey some prior knowledge to the pretrained language model to aid the report generation. Note that in RRG, we focus more on accuracy than diversity of the generated reports. We start by describing manual prompt learning with some designed manual prompts, and then introduce the details of the proposed automatic prompt learning.
\begin{table}
\caption{Illustrations of the manual prompts. `cn' in disease-enriched prompts refer to the class name.}
\vspace{-15pt}
\footnotesize
\begin{center}
\aboverulesep=0ex 
\belowrulesep=0ex 
\begin{tabular}{c|c|c}
\toprule
Knowledge &Label & Manual Prompt $\bm{p}$\\
\midrule
\multirow{2}{*}{common} &cmn1  & \textit{a picture that shows}   \\
\multirow{2}{*} &cmn2  & \textit{a picture of}   \\
\hline
\multirowcell{2}{domain\\ related} &dr1 & \textit{a radiology image that shows}  \\
\multirow{2}{*} &dr2 & \textit{a chest x-ray image that shows}  \\
\hline
\multirowcell{2}{disease\\ enriched} &de1 & \textit{a picture presenting [cn] that shows} \\
\multirow{2}{*} &de2 & \textit{a radiology image presenting [cn] that shows}\\
\bottomrule
\end{tabular}
\end{center}

\label{tab:manual_prompt}
\vspace{-20pt}
\end{table}

\subsubsection{Manual Prompt Learning}
Manual prompt learning requires prompt engineering to pre-define the prompts. In this work, we categorize the prompt designs according to different-levels of knowledge: common, domain-related and disease-enriched prompts as shown in~\autoref{tab:manual_prompt}. The common prompts, e.g. ``\textit{a picture that shows}", only contain the basic knowledge in natural scene image and tell the model that our purpose is to describe a picture. To add domain knowledge, we design some domain-related prompts including ``\textit{a radiology image that shows}" and ``\textit{a chest x-ray image that shows}" which informs the model that we are describing radiology images. 

Note that these manual prompts are the same for all the samples. Nonetheless, finding a manual prompt that benefits all samples is a great challenge, especially in RRG where the radiology image contains more fine-grained details and reports show sophisticated discourse relations. Hence, instance-level prompts with more medical information are required for clinically accurate radiology report generation. To this end, we propose an instance-level and disease-enriched manual prompt by introducing disease information of the sample into the prompt. The disease categories contain rich high-level information to guide the language model in encodingthe reports. 

Nevertheless, there are no ground truth category labels available in the commonly used RRG benchmark, i.e., MIMIC-CXR~\cite{johnson2019mimic}. As an alternative, we use the pseudo-labels on 14 common diseases in the chest X-ray (as shown in Table A1 in supplementary file) generated by an automatic labeller CheXpert~\cite{irvin2019chexpert} to form the manual prompt for each sample. Specifically, we insert the name of those classes showing presence into the non-class-related manual prompt as shown in the last two rows in~\autoref{tab:manual_prompt}. For example, an image with pseudo labels \underline{\textit{lung opacity, lung lesion}} is provided with a common and disease-related manual prompt: ``\textit{a picture presenting \textbf{lung opacity, lung lesion} that shows}". We also explore leveraging a LLM to generate the disease-enriched prompts \cite{zhou2022large}. Details and experimental results are given in Section A5 in supplementary file.

After setting the manual prompts $\bm{p}$ as listed in \autoref{tab:manual_prompt}, the next step is to form the prompt learning. In this paper, we prepend the prompt $\bm{p}$ to the report $\bm{R}$, hence the textual tokens sent to the embedding module changes to $\bm{R}^{\star} =\{\bm{p^{'}},\bm{R^{'}}\}$ where $\bm{p^{'}}$ as the tokenized manual prompt $\bm{p}$. $\bm{R}^{\star}$ is then fed into the remaining parts of the model performing the same operations as the report-only scenario. During the training, the whole model is trained by the loss:
\begin{align}
\label{eq:objective_manual}
\mathcal{L}_{ManCE}=-\sum_T \log P(y_T |f_{de}(\bm{L}^v,[\bm{L}^{mp},\bm{L}^{t}_{<T}]))
\end{align}
where $\bm{L}^{mp}$ are the prompt tokens output by the textual feature extractor.

\begin{table*}
\caption{Comparative results of our method with previous studies. The best values are highlighted in bold and the second best are underlined. 
}
\centering
\label{tab:main_results}
\begin{tabular}{cccccccc}
\toprule  
\textbf{Method} & \textbf{BLEU-1} & \textbf{BLEU-2} & \textbf{BLEU-3} &
\textbf{BLEU-4}  & \textbf{METEOR} &\textbf{CIDEr}\\
\midrule

 $ShowTell$ \cite{vinyals2015show} &30.8 &19.0 &12.5 &8.8  &12.2 & 9.6 \\
 $ATT2IN$ \cite{rennie2017self} &31.4 &19.8 &13.3 &9.5  & 12.2 &10.6\\
 $ADAATT$ \cite{lu2017knowing} &31.4 &19.8 &13.2 &9.4  & 12.8 &13.1\\
 $TopDown$ ~\cite{anderson2018bottom} &28.0 &16.9 &10.8 &7.4  & - &7.3 \\
  $PTSN$ ~\cite{zeng2022progressive} &27.0 &16.4 &11.0 &8.1  & 11.8&13.5 \\
$Transformer$~ \cite{vaswani2017attention} &31.6 &19.9 &14.0 &9.2 & 12.9 &13.4  \\
$M2Transformer$ \cite{cornia2020meshed} &33.2 &21.0 &14.2 &10.1 &13.4 &14.2 \\
 $CMCL$ \cite{liu-etal-2021-competence} &34.4 &21.7 &14.0 &9.7  & 13.3 & -\\

$R2Gen$ \cite{chen2020generating} &\textbf{35.3} &\underline{21.8} &{14.5} &10.3  & \underline{14.2} & 14.1\\ 
$R2GenCMN$ \cite{chen2021cross} &\textbf{35.3} &\underline{21.8} &\underline{14.8} &\underline{10.6}  &\underline{14.2} &14.3\\
$XPRONet$ \cite{wang2022cross} &34.4 &21.5 &{14.6} &{10.5}  & 13.8 &{15.4}\\

\midrule  
$\bm{Ours}-auto$  &32.4 &19.8 &13.4 &9.8    &13.3 &\underline{20.5}\\
$\bm{Ours}-manual(de1)$  &\underline{34.8} &\textbf{21.9} &\textbf{15.3} &\textbf{11.3} & \textbf{14.5} &\textbf{28.6}\\

\bottomrule 
\end{tabular}
\vspace{-10pt}
\end{table*}

\subsubsection{Automatic Prompt Learning}
\label{subsec:auto_prompt}
Manual prompt learning requires manual prompt engineering to design an appropriate prompt. Furthermore, the best-performing prompt is usually tailored for one dataset or one domain, hence a new round of manual prompt engineering may be required for a new domain. One may ask can we do this automatically without the need for great effort. In this paper, we give the answer that the model indeed has the capability of learning how to select the important information for prompting the pre-trained language model via a well-designed automatic prompt learning in RRG.

In particular, we randomly initialize an automatic prompt matrix $\bm{P}=\{p_1,p_2,...,p_{N^p}\}$ where $p_i \in \mathbb{R}^{1\times D_t}$ and $N^p$ refers to the number of automatic prompt tokens. Since $p_i$ has the same hidden size as a token embedding output by \autoref{eq:word_embed}, each element in $\bm{P}$ can be regarded as a prompt token. Then, the automatic prompt matrix is concatenated with the textual token embeddings from \autoref{eq:word_embed} and fed into the equations \ref{eq:pos_embed}\textasciitilde \ref{eq:dropln_embed} to enrich the position information and robustness. After adding the automatic prompt learning, the equation \ref{eq:word_embed} to \ref{eq:dropln_embed} changes to:
\begin{align}
\label{eq:word_embed_auto}
 o_i =&\ \bm{W}_{we}^T \cdot y_i,\quad \bm{P}=\{p_1,p_2,...,p_{N^p}\}, \\
 \label{eq:concat}
\bm{O}^{\prime}=&\ \{p_1,p_2,...,p_{N^p},o_1,o_2,...,o_{N^t}\}, \\
 \label{eq:pos_embed_auto}
 u_i =&\ \bm{W}_{pos}^T \cdot Pos(o^{\prime}_i), \quad {o^{\prime}}_i \in \bm{O^{\prime}}, \\
 \label{eq:dropln_embed_auto}
 e_i =&\  Dropout(LayerNorm({o^{\prime}}_i+u_i))
\end{align}

Accordingly, the loss to train the RRG model in \autoref{eq:objective_base} changes to:
\begin{align}
\label{eq:objective_auto}
\mathcal{L}_{AutoCE}=-\sum_T \log P(y_T |f_{de}(\bm{L}^v,[\bm{L}^{ap},\bm{L}^{t}_{<T}]))
\end{align}
where $\bm{L}^{ap}$ are the automatic prompt tokens output by the text encoder. Through equations \ref{eq:concat} to \ref{eq:dropln_embed_auto}, the automatic prompt enjoys the same position information and robustness as the textual tokens which plays a vital role in a transformer-based architecture. Moreover, instead of learning the whole embedding procedure (equations \ref{eq:word_embed} to \ref{eq:dropln_embed}), only requiring the automatic prompt matrix to learn a useful prompt token embedding (similar to \autoref{eq:word_embed}) is an easier task.

\section{Experiments}
\label{sec-experiments}

We verify the effectiveness of our proposed methods on the largest RRG benchmark, the MIMIC-CXR~\cite{johnson2019mimic} dataset, which is a recently released chest X-ray dataset with 377,110 images and 227,835 reports. This dataset is publicly available~\footnote{ https://physionet.org/content/MIMIC-cxr-jpg/2.0.0/}. The official data split is adopted here. Some statistics of this dataset are shown in Table A2 in the supplementary materials. Six widely used text generation evaluation metrics: BLEU\{1-4\}~\cite{papineni2002bleu}, 
METEOR~\cite{denkowski2011meteor} and CIDEr~\cite{vedantam2015cider} are utilized to assess the model performance.


\subsection{Implementation Details}
\label{subsec:implement}

We adopt the Swin-Transformer Base ~\cite{liu2021swin} pre-trained on ImageNet2K~\cite{deng2009imagenet} as our visual feature extractor. The first-three transformer layers of the pretrained Roberta~\cite{liu2019roberta} is employed as the text encoder. Note that the weights of the text encoder are frozen during training. The vision encoder and the decoder are the randomly initialized vanilla Transformer~\cite{vaswani2017attention} with $6$ and $3$ layers respectively, $8$ attention heads and $512$ dimensions for the hidden states. The length of the automatic prompt $N^p$ in~\autoref{eq:word_embed_auto} to~\autoref{eq:concat} is 8. Note that the optimal hyper-parameters are determined by evaluating the models on the validation sets. More implementation details, e.g., the data preprocessing and the optimizer settings, are described in Section A1 of the supplementary materials.

\subsection{Comparisons to SOTA methods}
\label{subsec:comparisons}
Here, we compare our method with $11$ state-of-the-art approaches including those widely used in image captioning \cite{vinyals2015show,rennie2017self,lu2017knowing,anderson2018bottom,cornia2020meshed,zeng2022progressive} and those designed for RRG \cite{liu-etal-2021-competence,chen2020generating,chen2021cross,wang2022cross}. As shown in~\autoref{tab:main_results}, our method with the disease-enriched manual prompt (label:de1) achieves the best performance on five out of six evaluation metrics (BLEU{2-4}, METEOR and CIDEr). The recent SOTA image captioning method PTSN fails to achieve promising results on RRG. The possible reason could be that it aims to capture the hierarchical semantic structure in textual space by clustering the words in the captions and then forming the prototype embedding. This process, however, is difficult to adapt to RRG since RRG is more fine-grained with more sophisticated discourse relationships and needs to focus at the sentence level rather than the word level to capture abnormal descriptions when generating the prototypes. It also can be seen that our method obtain slightly lower score than the best BLEU-1 score (-0.5\%). We hypothesize that this could be because of the enlarged vocabulary size and the presence of synonyms. BLEU-1 score computes the 1-gram word precision between the prediction and ground truth report. Nonetheless, the inclusion of the pre-trained language model enlarges the vocabulary size from $4,300$+ to $50,625$ and possesses a larger semantic word embedding space, hence the model may predict the synonym words or inflection which are not possible in previous studies. Some examples are shown in~\autoref{fig:example_vis} where our model with \textit{de1} prompt predicts \{\textit{heart}$\rightarrow$\textit{cardiac}\}, \{\textit{appears}$\rightarrow$\textit{re-demonstrated}\} and \{\textit{enlarged}$\rightarrow$\textit{enlargement}\}. Our model shows potent paraphrasing capability which leads to a lower score in some precise word-based evaluation metrics such as BLEU. Consequently, our method outperforms the previous studies on METEOR which considers synonyms in the metric computation. 

Note that our method (manual-de1) surpasses the previous methods by a large margin on the CIDEr score (+13.2), indicating that the model can better capture more key information in the generation. We mainly attribute this improvement to the disease information from the manual prompt which steers the models towards the related words in report generation. Our method with automatic prompts achieves comparable results as the commonly used image captioning methods, but is inferior to those methods specifically designed for the RRG. Nevertheless, automatic prompts boost the baseline model and demonstrates better performance than the common prompts and the domain-related prompts. We give the details of performance of all the manual prompts (as shown in~\autoref{tab:manual_prompt}) with different-levels of knowledge, the performance of the automatic prompt learning and a further discussion in the next section. Further qualitative results are shown in \autoref{fig:example_vis}.

\noindent \textbf{Use of Pseudo Labels} The disease-enriched manual prompts incorporate pseudo labels. Note that our visual extractor is IMAGENET2K pre-trained. Method $CMCL$~\cite{liu-etal-2021-competence} uses the pseudo labels to fine-tune their visual extractor and determine the visual difficulty; XPRONet~\cite{wang2022cross} utilizes the pseudo labels to form a cross-modal prototype-driven network. Our method still remarkably outperforms these methods. Moreover, the pseudo labels we use are those provided with the MIMIC-CXR dataset.

\begin{table*}
\caption{The experimental results of ablation studies on the MIMIC-CXR datasets. The best values are highlighted in bold. 
}
\centering
\label{tab:ablation_studies}
\begin{tabular}{c|c|cccccccc}

\toprule  
\textbf{Knowledge} &\textbf{Methods}  & \textbf{BLEU-1} & \textbf{BLEU-2} & \textbf{BLEU-3} &
\textbf{BLEU-4} & \textbf{METEOR} & \textbf{CIDEr}  \\
\midrule  

- &Base  &31.7 &19.3 &13.0 &9.4 &13.0 & 19.5 \\
\hline
\multirow{2}{*}{common} &+ cmn1 &31.5 &19.1 &13.1 &9.6 &13.0 &18.4   \\
\multirow{2}{*} &+ cmn2  &31.9  &19.3 &13.1 &9.5 &13.1 & 19.9  \\
\hline
\multirowcell{2}{domain\\ related} &+ dr1  &31.8 &19.2 &13.1 &9.6 &13.0 & 21.3 \\
\multirow{2}{*} &+ dr2  &31.1 &18.9 &12.7 &9.2 &12.9 & 17.4 \\
\hline
\multirowcell{2}{disease\\ enriched} &+ de2  &34.6 &21.7 &15.1 &\textbf{11.3} &14.4 & 28.1 \\
\multirow{2}{*} &+ de1  &\textbf{34.8} &\textbf{21.9} &\textbf{15.3} &\textbf{11.3} & \textbf{14.5} &\textbf{28.6} \\
\hline
\multirow{2}{*}{Auto} &+ all &31.7 &19.2 &13.0 &9.5 &12.9 &19.6 \\
\multirow{2}{*} &+ word &32.4 &19.8 &13.4 &9.8 &13.3 &20.5 \\


\bottomrule 
\end{tabular}
\vspace{-10pt}
\end{table*}

\subsection{Ablation Studies}
\label{subsec:ablation_studies}
\subsubsection{Effect of Different Manual Prompts}
We conduct ablation studies to further verify the effectiveness of our proposed methods and investigate the performance of different types of prompts. As shown in~\autoref{tab:ablation_studies}, the lowest knowledge-level prompts, i.e., common prompts, showing effectiveness in various vision-language tasks fail to bring notable improvement compared with the baseline model, demonstrating the difficulty in prompting the model in this trickier RRG task. Further adding domain-related knowledge into the common prompt, i.e. ``\textit{radiology image}" (dr1), shows slight improvements, while a notable performance drop can be seen with more task-specific domain-related prompt, i.e., ``\textit{chest x-ray image}". We conjecture this is caused by the knowledge being missing in the pretrained language model Roberta, which is mainly trained on several natural scene corpus such as news, stories and English Wikipedia. Therefore, the pretrained text encoder might contain some knowledge about a relatively larger domain, i.e. radiology images, but fails in eliciting knowledge to the more specific domain of chest of X-rays that seldom occur in the training corpus. We also experimented with two medical text encoders~\cite{alsentzer2019publicly,boecking2022making}, but this is shown to be less coherent and cohesive than Roberta when generating reports (experimental results are given in Section A5 in the supplementary file), despite containing more medical information. From the above results, we might conclude that leveraging pretrained language models with prompt learning, simply increasing domain knowledge cannot demonstrate notably superior performance than the common prompts since it is of great importance to strike a balance between the better encoding capability normally shown in a natural scene text encoder with common prompts and providing more medical information for the clinical text encoder or domain-related prompts.

Our model with disease-enriched manual prompts achieves the best performance among three types of prompts and notably improves the baseline model on all evaluation metrics as the model seems to acquire more useful knowledge for each sample from the instance-level and disease-enriched prompts. In detail, \textbf{de2} prompt shows slightly better results than \textbf{de1}. This performance gain possibly comes from the combination of different types of knowledge since \textbf{de2} combines the common and medical disease knowledge, while \textbf{de2} only contains medical knowledge. 

Overall, the above results verify the effectiveness of our proposed methods and demonstrate that leveraging prompt learning with the PLM to distil the knowledge can elegantly achieve better results without sophisticated module designs, e.g. new loss function and architecture changes. The performance and details of more different types of prompts, leveraging a large language model as the prompt engineer, different language models as text encoders, use of a visual-language model, and the associated analysis are given in the supplementary files.

\subsubsection{Effect of Automatic Prompts}
We evaluate the performance of the proposed automatic prompt, shown in the last two rows in~\autoref{tab:ablation_studies}. We compare the two implementations of the automatic prompt learning: (1) \textbf{Auto+all} means that the automatic prompt matrix learn the whole embedding procedure, i.e. \autoref{eq:word_embed_auto} to \autoref{eq:dropln_embed_auto}; (2)~\textbf{Auto+word} only expects the the automatic prompt matrix to learn the word embedding \autoref{eq:word_embed_auto}. As the results show, automatic prompt learning the word embedding shows notably better performance than the one learning the whole embedding procedure, which proves our assumption that expecting the model to learn the whole embedding procedures is too difficult and the importance of the position information. Moreover, automatic prompts (word) outperform the common and domain-related prompts and modestly improves the baseline model on all the evaluation metrics. Note that the MIMIC-CXR dataset is very large and challenging and making even a small improvement proves difficult. Although there is still a significant performance gap between automatic prompts and disease-enriched prompts, automatic prompts are formed without complicated manual prompt engineering and extra input sources. A visual comparison to the baseline model is shown in \autoref{sec:qualitative_results} to further demonstrate the potential of the automatic prompt learning.

\begin{figure}[b]
    \vspace{-15pt}
    \centering 
    \includegraphics[width=0.4\textwidth]{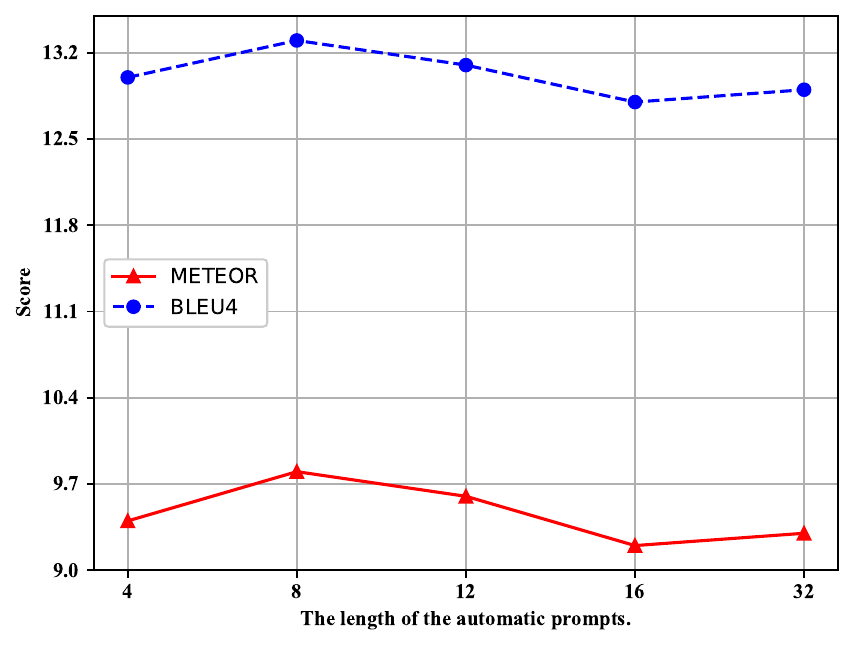} 
    \caption{Effect of varying the length of the automatic prompt.}    
    \label{fig:length_ablation_iu}  
    \vspace{-15pt}
\end{figure}

\begin{figure*}
    \centering 
    \includegraphics[width=0.89\textwidth]{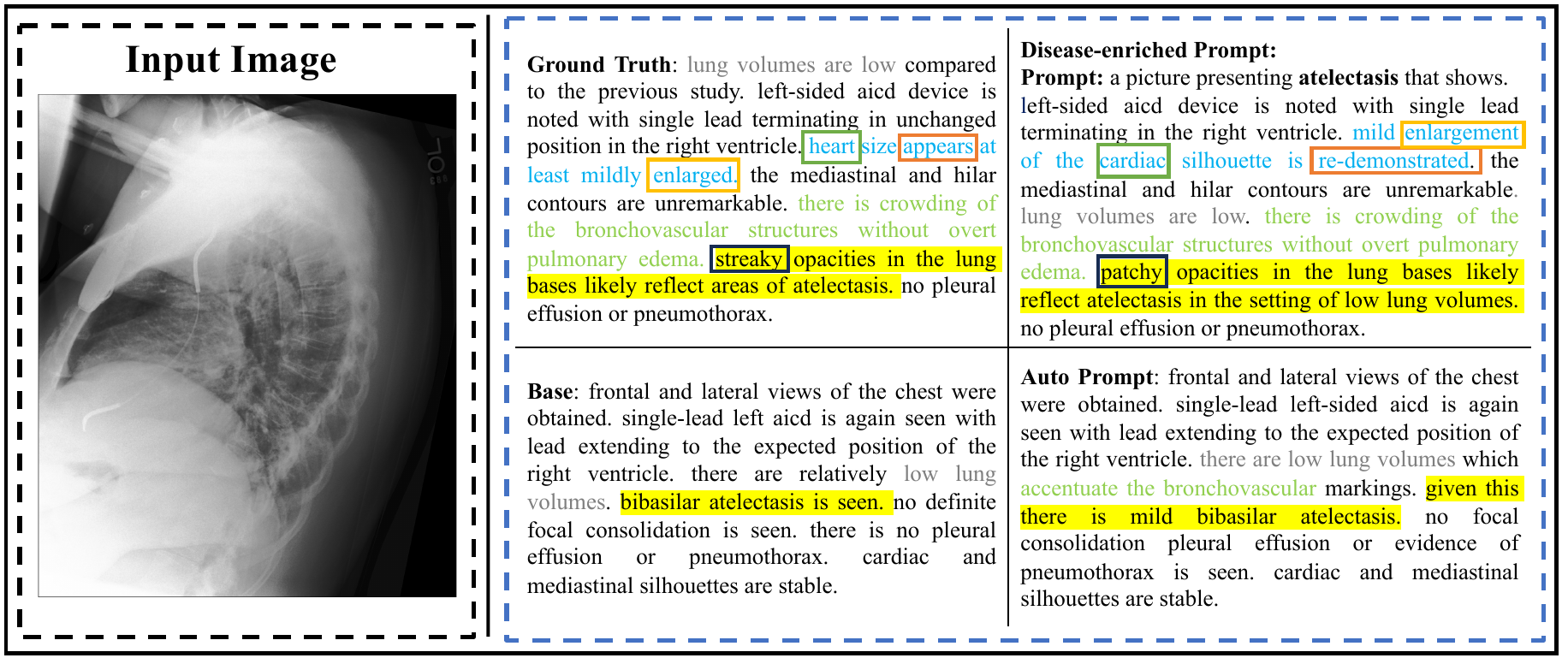} 
    \caption{An example of the reports generated by different models. Texts describing the same abnormalities are highlighted in same colors. Words with the same colored rectangular represent the synonym between the ground truth and the prediction by \textit{de1} prompt. }    
    \label{fig:example_vis}  
\vspace{-15pt}
\end{figure*}

\noindent \textbf{The influence of the Length in APL} We conduct an ablation study to explore the influence of the only hyperparameter, the length of the prompt $N^p$, in the proposed automatic prompt learning by varying $N^p$ from $4$ to $32$. As illustrated in \autoref{fig:length_ablation_iu}, the automatic prompt learning is not overly sensitive to length of the prompts.  However, it is still important to strike a balance when setting the value of $N^p$ since a shorter prompt may not incorporate all the useful information, while too large a value may introduce noisy information. For example, reducing the $N^p$ from $8$ to $4$ or increasing it from $8$ to $32$ both steadily lead to a performance drop.

\subsection{Qualitative Results}
\label{sec:qualitative_results}

To further verify the effectiveness of the proposed method, we provide an example in \autoref{fig:example_vis} which shows the reports generated by the \textbf{Base} model and models with different prompts. Texts that describe the same abnormalities have the same color and synonym words, and inflections are highlighted by similar colored rectangles. The disease-enriched prompt can not only generate clearly more accurate normal descriptions, but also show the potent capability of capturing abnormalities too. For example, assisted by the disease knowledge from the prompt, i.e., ``\textit{a picture presenting \textbf{atelectasis} that shows.}", our model can generate a more concrete description for the \textbf{atelectasis} (marked by yellow color) compared with the simple description ``\textit{bibasilar atelectasis is seen.}". Moreover, we also found an interesting observation that the model gives the supporting statements for abnormalities diagnosis when adding the prompt. For instance, sentence ``\textit{patchy opacities in the lung bases likely reflect atelectasis \textbf{in the setting of low lung volumes.}} explicitly gives the condition for generating the findings for \textbf{atelectasis}. The added disease information triggers the model to consider the possible causes, which may explain why the disease-enriched prompt can accurately capture other abnormalities ``\textit{enlarged heart size}" and ``\textit{crowding of the bronchovascular}" while \textbf{Base} model produces ``\textit{cardiac and mediastinal silhouettes are stable.}".

A similar pattern can be seen in automatic prompt where the model can capture the ``\textit{accentuate the bronchovascular}" and conclude the ``\textit{mild bibasilar atelectasis}" from the condition of lung volume in the previous sentence, indicating that the automatic prompt genuinely records some useful knowledge. Although the automatic prompt shows better performance and captures more details than the \textbf{Base} model, there is still a gap between the automatic prompt and disease-enriched prompt both in generating the normal and abnormal descriptions due to its having less medical knowledge. Nonetheless, the automatic prompt relieves having to depend on rigorous manual prompt engineering and extra input sources.  We believe that it is worth exploring automatic prompt learning on more challenging tasks and non-natural scene scenarios which require experts to design useful prompts. Finally, this example also shows that the inclusion of a pretrained encoder gives our model powerful paraphrasing capabilities. This may result in lower scores on evaluation metrics, but the generated reports seem to be clinically accurate and have greater diversity.

\section{Conclusions}
\label{sec:conclusion}
In this paper, we present PromptRRG, a new framework for radiology report generation that focuses on distilling prior domain knowledge via prompt learning. Unlike previous works, we propose to leverage the existing knowledge of a pretrained language model and carefully design some manual prompts with different-levels of knowledge. To alleviate the need for rigorous manual prompt engineering, we propose an automatic prompt learning approach which renders the model to learn and record the most useful prompts. Experimental results demonstrate that the proposed disease-enriched prompts can elegantly distill important medical knowledge without introducing trainable parameters and achieve SOTA performance on the MIMIC-CXR dataset. Ablation studies also prove the effectiveness of the proposed automatic prompt learning.

\clearpage
{\small
\bibliographystyle{ieee_fullname}
\bibliography{egbib}

\begin{thebibliography}{10}\itemsep=-1pt

\bibitem{NEURIPS2022_960a172b}
Jean-Baptiste Alayrac, Jeff Donahue, Pauline Luc, Antoine Miech, Iain Barr,
  Yana Hasson, Karel Lenc, Arthur Mensch, Katherine Millican, Malcolm Reynolds,
  Roman Ring, Eliza Rutherford, Serkan Cabi, Tengda Han, Zhitao Gong, Sina
  Samangooei, Marianne Monteiro, Jacob~L Menick, Sebastian Borgeaud, Andy
  Brock, Aida Nematzadeh, Sahand Sharifzadeh, Miko\l~aj Bi\'{n}kowski, Ricardo
  Barreira, Oriol Vinyals, Andrew Zisserman, and Kar\'{e}n Simonyan.
\newblock Flamingo: a visual language model for few-shot learning.
\newblock In S. Koyejo, S. Mohamed, A. Agarwal, D. Belgrave, K. Cho, and A. Oh,
  editors, {\em Advances in Neural Information Processing Systems}, volume~35,
  pages 23716--23736. Curran Associates, Inc., 2022.

\bibitem{alsentzer2019publicly}
Emily Alsentzer, John~R Murphy, Willie Boag, Wei-Hung Weng, Di Jin, Tristan
  Naumann, WA Redmond, and Matthew~BA McDermott.
\newblock Publicly available clinical bert embeddings.
\newblock {\em NAACL HLT 2019}, page~72, 2019.

\bibitem{anderson2018bottom}
Peter Anderson, Xiaodong He, Chris Buehler, Damien Teney, Mark Johnson, Stephen
  Gould, and Lei Zhang.
\newblock Bottom-up and top-down attention for image captioning and visual
  question answering.
\newblock In {\em Proceedings of the IEEE Conference on Computer Vision and
  Pattern Recognition}, pages 6077--6086, 2018.

\bibitem{bach-etal-2022-promptsource}
Stephen Bach, Victor Sanh, Zheng~Xin Yong, Albert Webson, Colin Raffel,
  Nihal~V. Nayak, Abheesht Sharma, Taewoon Kim, M~Saiful Bari, Thibault Fevry,
  Zaid Alyafeai, Manan Dey, Andrea Santilli, Zhiqing Sun, Srulik Ben-david,
  Canwen Xu, Gunjan Chhablani, Han Wang, Jason Fries, Maged Al-shaibani, Shanya
  Sharma, Urmish Thakker, Khalid Almubarak, Xiangru Tang, Dragomir Radev, Mike
  Tian-jian Jiang, and Alexander Rush.
\newblock {P}rompt{S}ource: An integrated development environment and
  repository for natural language prompts.
\newblock In {\em Proceedings of the 60th Annual Meeting of the Association for
  Computational Linguistics: System Demonstrations}, pages 93--104, Dublin,
  Ireland, May 2022. Association for Computational Linguistics.

\bibitem{boecking2022making}
Benedikt Boecking, Naoto Usuyama, Shruthi Bannur, Daniel~C Castro, Anton
  Schwaighofer, Stephanie Hyland, Maria Wetscherek, Tristan Naumann, Aditya
  Nori, Javier Alvarez-Valle, et~al.
\newblock Making the most of text semantics to improve biomedical
  vision--language processing.
\newblock In {\em European conference on computer vision}, pages 1--21.
  Springer, 2022.

\bibitem{brown2020language}
Tom Brown, Benjamin Mann, Nick Ryder, Melanie Subbiah, Jared~D Kaplan, Prafulla
  Dhariwal, Arvind Neelakantan, Pranav Shyam, Girish Sastry, Amanda Askell,
  Sandhini Agarwal, Ariel Herbert-Voss, Gretchen Krueger, Tom Henighan, Rewon
  Child, Aditya Ramesh, Daniel Ziegler, Jeffrey Wu, Clemens Winter, Chris
  Hesse, Mark Chen, Eric Sigler, Mateusz Litwin, Scott Gray, Benjamin Chess,
  Jack Clark, Christopher Berner, Sam McCandlish, Alec Radford, Ilya Sutskever,
  and Dario Amodei.
\newblock Language models are few-shot learners.
\newblock In H. Larochelle, M. Ranzato, R. Hadsell, M.F. Balcan, and H. Lin,
  editors, {\em Advances in Neural Information Processing Systems}, volume~33,
  pages 1877--1901. Curran Associates, Inc., 2020.

\bibitem{chen2021cross}
Zhihong Chen, Yaling Shen, Yan Song, and Xiang Wan.
\newblock Cross-modal memory networks for radiology report generation.
\newblock In {\em Proceedings of the 59th Annual Meeting of the Association for
  Computational Linguistics and the 11th International Joint Conference on
  Natural Language Processing (Volume 1: Long Papers)}, pages 5904--5914, 2021.

\bibitem{chen2020generating}
Zhihong Chen, Yan Song, Tsung-Hui Chang, and Xiang Wan.
\newblock Generating radiology reports via memory-driven transformer.
\newblock In {\em Proceedings of the 2020 Conference on Empirical Methods in
  Natural Language Processing (EMNLP)}, pages 1439--1449, 2020.

\bibitem{cornia2020meshed}
Marcella Cornia, Matteo Stefanini, Lorenzo Baraldi, and Rita Cucchiara.
\newblock Meshed-memory transformer for image captioning.
\newblock In {\em Proceedings of the IEEE Conference on Computer Vision and
  Pattern Recognition}, pages 10578--10587, 2020.

\bibitem{deng2009imagenet}
Jia Deng, Wei Dong, Richard Socher, Li-Jia Li, Kai Li, and Li Fei-Fei.
\newblock Imagenet: A large-scale hierarchical image database.
\newblock In {\em 2009 IEEE Conference on Computer Vision and Pattern
  Recognition}, pages 248--255. Ieee, 2009.

\bibitem{denkowski2011meteor}
Michael Denkowski and Alon Lavie.
\newblock Meteor 1.3: Automatic metric for reliable optimization and evaluation
  of machine translation systems.
\newblock In {\em Proceedings of the sixth workshop on Statistical Machine
  Translation}, pages 85--91, 2011.

\bibitem{dou2022empirical}
Zi-Yi Dou, Yichong Xu, Zhe Gan, Jianfeng Wang, Shuohang Wang, Lijuan Wang,
  Chenguang Zhu, Pengchuan Zhang, Lu Yuan, Nanyun Peng, et~al.
\newblock An empirical study of training end-to-end vision-and-language
  transformers.
\newblock In {\em Proceedings of the IEEE Conference on Computer Vision and
  Pattern Recognition}, pages 18166--18176, 2022.

\bibitem{guo2020normalized}
Longteng Guo, Jing Liu, Xinxin Zhu, Peng Yao, Shichen Lu, and Hanqing Lu.
\newblock Normalized and geometry-aware self-attention network for image
  captioning.
\newblock In {\em Proceedings of the IEEE Conference on Computer Vision and
  Pattern Recognition}, pages 10327--10336, 2020.

\bibitem{he2016deep}
Kaiming He, Xiangyu Zhang, Shaoqing Ren, and Jian Sun.
\newblock Deep residual learning for image recognition.
\newblock In {\em Proceedings of the IEEE Conference on Computer Vision and
  Pattern Recognition}, pages 770--778, 2016.

\bibitem{irvin2019chexpert}
Jeremy Irvin, Pranav Rajpurkar, Michael Ko, Yifan Yu, Silviana Ciurea-Ilcus,
  Chris Chute, Henrik Marklund, Behzad Haghgoo, Robyn Ball, Katie Shpanskaya,
  et~al.
\newblock Chexpert: A large chest radiograph dataset with uncertainty labels
  and expert comparison.
\newblock In {\em Proceedings of the AAAI conference on artificial
  intelligence}, volume~33, pages 590--597, 2019.

\bibitem{ji2021improving}
Jiayi Ji, Yunpeng Luo, Xiaoshuai Sun, Fuhai Chen, Gen Luo, Yongjian Wu, Yue
  Gao, and Rongrong Ji.
\newblock Improving image captioning by leveraging intra-and inter-layer global
  representation in transformer network.
\newblock In {\em Proceedings of the AAAI Conference on Artificial
  Intelligence}, volume~35, pages 1655--1663, 2021.

\bibitem{jin2022good}
Woojeong Jin, Yu Cheng, Yelong Shen, Weizhu Chen, and Xiang Ren.
\newblock A good prompt is worth millions of parameters: Low-resource
  prompt-based learning for vision-language models.
\newblock In {\em Proceedings of the 60th Annual Meeting of the Association for
  Computational Linguistics (Volume 1: Long Papers)}, pages 2763--2775, 2022.

\bibitem{johnson2019mimic}
Alistair~EW Johnson, Tom~J Pollard, Nathaniel~R Greenbaum, Matthew~P Lungren,
  Chih-ying Deng, Yifan Peng, Zhiyong Lu, Roger~G Mark, Seth~J Berkowitz, and
  Steven Horng.
\newblock Mimic-cxr-jpg, a large publicly available database of labeled chest
  radiographs.
\newblock {\em arXiv preprint arXiv:1901.07042}, 2019.

\bibitem{kale2023kgvl}
Kaveri Kale, Pushpak Bhattacharyya, Milind Gune, Aditya Shetty, and Rustom
  Lawyer.
\newblock Kgvl-bart: Knowledge graph augmented visual language bart for
  radiology report generation.
\newblock In {\em Proceedings of the 17th Conference of the European Chapter of
  the Association for Computational Linguistics}, pages 3393--3403, 2023.

\bibitem{kenton2019bert}
Jacob Devlin Ming-Wei~Chang Kenton and Lee~Kristina Toutanova.
\newblock Bert: Pre-training of deep bidirectional transformers for language
  understanding.
\newblock In {\em Proceedings of NAACL-HLT}, pages 4171--4186, 2019.

\bibitem{li2019knowledge}
Christy~Y Li, Xiaodan Liang, Zhiting Hu, and Eric~P Xing.
\newblock Knowledge-driven encode, retrieve, paraphrase for medical image
  report generation.
\newblock In {\em Proceedings of the AAAI Conference on Artificial
  Intelligence}, volume~33, pages 6666--6673, 2019.

\bibitem{li2022blip}
Junnan Li, Dongxu Li, Caiming Xiong, and Steven Hoi.
\newblock Blip: Bootstrapping language-image pre-training for unified
  vision-language understanding and generation.
\newblock In {\em International Conference on Machine Learning}, pages
  12888--12900. PMLR, 2022.

\bibitem{lin2014microsoft}
Tsung-Yi Lin, Michael Maire, Serge Belongie, James Hays, Pietro Perona, Deva
  Ramanan, Piotr Doll{\'a}r, and C~Lawrence Zitnick.
\newblock Microsoft coco: Common objects in context.
\newblock In {\em Computer Vision--ECCV 2014: 13th European Conference, Zurich,
  Switzerland, September 6-12, 2014, Proceedings, Part V 13}, pages 740--755.
  Springer, 2014.

\bibitem{liu-etal-2021-competence}
Fenglin Liu, Shen Ge, and Xian Wu.
\newblock Competence-based multimodal curriculum learning for medical report
  generation.
\newblock In {\em Proceedings of the 59th Annual Meeting of the Association for
  Computational Linguistics and the 11th International Joint Conference on
  Natural Language Processing (Volume 1: Long Papers)}, pages 3001--3012,
  Online, 2021. Association for Computational Linguistics.

\bibitem{liu2021exploring}
Fenglin Liu, Xian Wu, Shen Ge, Wei Fan, and Yuexian Zou.
\newblock Exploring and distilling posterior and prior knowledge for radiology
  report generation.
\newblock In {\em Proceedings of the IEEE Conference on Computer Vision and
  Pattern Recognition}, pages 13753--13762, 2021.

\bibitem{liu2021contrastive}
Fenglin Liu, Changchang Yin, Xian Wu, Shen Ge, Ping Zhang, and Xu Sun.
\newblock Contrastive attention for automatic chest x-ray report generation.
\newblock In {\em Findings of the Association for Computational Linguistics:
  ACL-IJCNLP 2021}, pages 269--280, 2021.

\bibitem{liu2019clinically}
Guanxiong Liu, Tzu-Ming~Harry Hsu, Matthew McDermott, Willie Boag, Wei-Hung
  Weng, Peter Szolovits, and Marzyeh Ghassemi.
\newblock Clinically accurate chest x-ray report generation.
\newblock In {\em Machine Learning for Healthcare Conference}, pages 249--269.
  PMLR, 2019.

\bibitem{liu2019roberta}
Yinhan Liu, Myle Ott, Naman Goyal, Jingfei Du, Mandar Joshi, Danqi Chen, Omer
  Levy, Mike Lewis, Luke Zettlemoyer, and Veselin Stoyanov.
\newblock Roberta: A robustly optimized bert pretraining approach.
\newblock {\em arXiv preprint arXiv:1907.11692}, 2019.

\bibitem{liu2021swin}
Ze Liu, Yutong Lin, Yue Cao, Han Hu, Yixuan Wei, Zheng Zhang, Stephen Lin, and
  Baining Guo.
\newblock Swin transformer: Hierarchical vision transformer using shifted
  windows.
\newblock In {\em Proceedings of the IEEE international Conference on Computer
  Vision}, pages 10012--10022, 2021.

\bibitem{lu2017knowing}
Jiasen Lu, Caiming Xiong, Devi Parikh, and Richard Socher.
\newblock Knowing when to look: Adaptive attention via a visual sentinel for
  image captioning.
\newblock In {\em Proceedings of the IEEE Conference on Computer Vision and
  Pattern Recognition}, pages 375--383, 2017.

\bibitem{lu2022prompt}
Yuning Lu, Jianzhuang Liu, Yonggang Zhang, Yajing Liu, and Xinmei Tian.
\newblock Prompt distribution learning.
\newblock In {\em Proceedings of the IEEE/CVF Conference on Computer Vision and
  Pattern Recognition}, pages 5206--5215, 2022.

\bibitem{pan2020x}
Yingwei Pan, Ting Yao, Yehao Li, and Tao Mei.
\newblock X-linear attention networks for image captioning.
\newblock In {\em Proceedings of the IEEE Conference on Computer Vision and
  Pattern Recognition}, pages 10971--10980, 2020.

\bibitem{papineni2002bleu}
Kishore Papineni, Salim Roukos, Todd Ward, and Wei-Jing Zhu.
\newblock Bleu: a method for automatic evaluation of machine translation.
\newblock In {\em Proceedings of the 40th annual meeting of the Association for
  Computational Linguistics}, pages 311--318, 2002.

\bibitem{rennie2017self}
Steven~J Rennie, Etienne Marcheret, Youssef Mroueh, Jerret Ross, and Vaibhava
  Goel.
\newblock Self-critical sequence training for image captioning.
\newblock In {\em Proceedings of the IEEE Conference on Computer Vision and
  Pattern Recognition}, pages 7008--7024, 2017.

\bibitem{LLaMAnotarxivyet}
Hugo Touvron, Thibaut Lavril, Gautier Izacard, Xavier Martinet, Marie-Anne
  Lachaux, Timothee Lacroix, Baptiste Rozière, Naman Goyal, Eric Hambro,
  Faisal Azhar, Aurelien Rodriguez, Armand Joulin, Edouard Grave, and
  Guillaume. Lample.
\newblock Llama: Open and efficient foundation language models, 2023.

\bibitem{vaswani2017attention}
Ashish Vaswani, Noam Shazeer, Niki Parmar, Jakob Uszkoreit, Llion Jones,
  Aidan~N Gomez, {\L}ukasz Kaiser, and Illia Polosukhin.
\newblock Attention is all you need.
\newblock {\em Advances in Neural Information Processing Systems}, 30, 2017.

\bibitem{vedantam2015cider}
Ramakrishna Vedantam, C Lawrence~Zitnick, and Devi Parikh.
\newblock Cider: Consensus-based image description evaluation.
\newblock In {\em Proceedings of the IEEE Conference on Computer Vision and
  Pattern Recognition}, pages 4566--4575, 2015.

\bibitem{vinyals2015show}
Oriol Vinyals, Alexander Toshev, Samy Bengio, and Dumitru Erhan.
\newblock Show and tell: A neural image caption generator.
\newblock In {\em Proceedings of the IEEE Conference on Computer Vision and
  Pattern Recognition}, pages 3156--3164, 2015.

\bibitem{wang2022cross}
Jun Wang, Abhir Bhalerao, and Yulan He.
\newblock Cross-modal prototype driven network for radiology report generation.
\newblock In {\em Computer Vision--ECCV 2022: 17th European Conference, Tel
  Aviv, Israel, October 23--27, 2022, Proceedings, Part XXXV}, pages 563--579.
  Springer, 2022.

\bibitem{wang2022controllable}
Ning Wang, Jiahao Xie, Jihao Wu, Mingbo Jia, and Linlin Li.
\newblock Controllable image captioning via prompting.
\newblock In {\em Proceedings of the AAAI Conference on Artificial
  Intelligence}, 2023.

\bibitem{wang2021simvlm}
Zirui Wang, Jiahui Yu, Adams~Wei Yu, Zihang Dai, Yulia Tsvetkov, and Yuan Cao.
\newblock Simvlm: Simple visual language model pretraining with weak
  supervision.
\newblock {\em arXiv preprint arXiv:2108.10904}, 2021.

\bibitem{zeng2022progressive}
Pengpeng Zeng, Jinkuan Zhu, Jingkuan Song, and Lianli Gao.
\newblock Progressive tree-structured prototype network for end-to-end image
  captioning.
\newblock In {\em Proceedings of the 30th ACM International Conference on
  Multimedia}, pages 5210--5218, 2022.

\bibitem{zhang2020radiology}
Yixiao Zhang, Xiaosong Wang, Ziyue Xu, Qihang Yu, Alan Yuille, and Daguang Xu.
\newblock When radiology report generation meets knowledge graph.
\newblock In {\em Proceedings of the AAAI Conference on Artificial
  Intelligence}, volume~34, pages 12910--12917, 2020.

\bibitem{zhou2022learning}
Kaiyang Zhou, Jingkang Yang, Chen~Change Loy, and Ziwei Liu.
\newblock Learning to prompt for vision-language models.
\newblock {\em International Journal of Computer Vision}, 130(9):2337--2348,
  2022.

\bibitem{zhou2022large}
Yongchao Zhou, Andrei~Ioan Muresanu, Ziwen Han, Keiran Paster, Silviu Pitis,
  Harris Chan, and Jimmy Ba.
\newblock Large language models are human-level prompt engineers.
\newblock In {\em The Eleventh International Conference on Learning
  Representations}, 2022.

\bibitem{zhou2023large}
Yongchao Zhou, Andrei~Ioan Muresanu, Ziwen Han, Keiran Paster, Silviu Pitis,
  Harris Chan, and Jimmy Ba.
\newblock Large language models are human-level prompt engineers.
\newblock In {\em The Eleventh International Conference on Learning
  Representations}, 2023.

\end{thebibliography}
}
\end{document}




\title{Can Prompt Learning Benefit Radiology Report Generation?\\
(Supplementary Material)} 

%


\author{Jun Wang$^1$, Lixing Zhu$^{2,3}$, Abhir Bhalerao$^1$ and Yulan He$^{1,2,3}$}
\institute{$^1$University of Warwick, UK; $^2$King's College London, UK \\
$^3$The Alan Turing Institute, UK\\}

\maketitle

\setcounter{table}{0}
\renewcommand{\thetable}{A\arabic{table}}
\setcounter{figure}{0}
\renewcommand{\thefigure}{A\arabic{figure}}

\section{Further Implementation Details}
Images are first cropped (resized) to $(224, 224)$ during the training (and inference) phase. In addition, we apply a random horizontal flip to further augment the datasets during training. The downsampling rate of the visual extractor is $32$, hence there are $N^s=7\times7=49$ visual tokens. The same as the previous works~\cite{li2018hybrid,chen2020generating,chen2021cross,wang2022cross}. We pre-process the reports by converting all tokens to lower cases and removing non-alphanumeric characters.

AdamW~\cite{loshchilovdecoupled} is used to optimize the whole model. The learning rates are set to $5e-5$ for the visual feature extractor and vision encoder, and $2.5e-4$ for the decoder and $5e-4$ for the report generation head. We train our model with the batch size of $64$. Note that the optimal hyper-parameters are determined by evaluating the models on the validation sets. 

Same as the recent state-of-the-art studies \cite{li2018hybrid,li2019knowledge,chen2020generating,chen2021cross}, we adopt Beam Search as the sampling method during the inference phase. The beam size is set to 3 to balance the effectiveness and efficiency. We implement our model via the PyTorch~\cite{paszke2019pytorch} deep learning framework on Nvidia A100 GPU cards. 

\section{14 Categories of Chest X-Ray Observations}
\label{sec:observation}
\autoref{tab:categories} shows the 14 categories for the multi-label classification set by the automatic labeller.

\begin{table}
\normalsize
\caption{The 14 categories predefined and output by the automatic labeler.}
\begin{center}
\begin{tabular}{|c|c|c|}
\hline
 Lung Opacity& Cardiomegaly & No Finding \\
\hline
Lung Lesion &Consolidation  &Edema   \\
\hline
 Pneumothorax  &Pneumonia &Atelectasis  \\
\hline 
Pleural Effusion &Pleural Other & Fracture \\
\hline
Support Devices
&\multicolumn{2}{|c|}{Enlarged Cardiomediastinum}  \\
\hline
\end{tabular}
\end{center}
\label{tab:categories}
\end{table}

\section{Some Statistics of MIMIC-CXR}
\autoref{tab:dataset} gives some statistics of MIMIC-CXR including the number of samples in each subset and the average report length. Note that some images are associated with the same report, e.g., same patient with different views (front or lateral). Samples without a radiology report are excluded. 

\begin{table}
\normalsize
\caption{Some statistics of MIMIC-CXR dataset: \# refers the number of; Avg.Len denotes the average length of the reports.}
\begin{center}
\begin{tabular}{|c|c|c|c|c|}
\hline
 \textbf{Subset} &\textbf{ \#Image } &\textbf{ \#Report } &\textbf{ \#Sample } &\textbf{ Avg.Len } \\
\hline
Train &368,960 &222,758 &270,790 &53.20   \\
\hline
Validation &2,991 &1,808 &2,130 &53.28  \\
\hline
Test &5,159 &3,269 &3,858 &66.76  \\
\hline
\end{tabular}
\end{center}
\label{tab:dataset}
\vspace{-20pt}
\end{table}

\begin{table}
\normalsize
\caption{Illustrations of the manual prompts.}
\begin{center}
\aboverulesep=0ex 
\belowrulesep=0ex 
\begin{tabular}{c|c|c}
\toprule
\textbf{ Knowledge } &\textbf{ Label } & \textbf{Manual Prompt $\bm{p}$}\\
\midrule
\multirow{2}{*}{common} &cmn3  & \textit{a photo that shows}   \\
\multirow{2}{*} &cmn4  & \textit{a photo of}   \\
\hline
\multirowcell{2}{domain\\ related} &dr3 & \textit{a radiology image of} \\
\multirow{2}{*} &dr4 & \textit{a chest x-ray image of}  \\
\hline
\multirowcell{4}{disease\\ enriched} &de3 & \textit{a picture of [class name]}  \\
\multirow{4}{*} &de4 & \textit{a/an [class name] picture}  \\
\multirow{4}{*} &de5 & \textit{a/an [class name] picture that shows}  \\
\multirow{4}{*} &de6 & \textit{ a photo presenting [class name] that shows} \\
\multirow{4}{*} &de7 & \textit{ a chest x-ray image presenting [class name] that shows} \\
\bottomrule
\end{tabular}
\end{center}
\label{tab:manual_prompt}
\vspace{-30pt}
\end{table}

\section{The Performance of More Manual Prompts}
Here, we demonstrate the performance of more diverse manual prompts to inspire the future investigation. The manual prompts are shown in \autoref{tab:manual_prompt} and the associated performance is given in \autoref{tab:prompt_results}. 

As shown in \autoref{tab:prompt_results} and Table 3 in the main paper manuscript, the model is relatively sensitive to the design of the prompts, especially for the diseased-enriched prompts where a larger performance variance can seen than the common and domain-related prompts. We conjecture that this is because the disease-enriched prompts are specified for each sample (instance-level), hence there are numbers of prompt instances for each disease-enriched prompt template compared with the only one instance (itself) in the common and domain-related prompts. To be more specific, the [class name] in the disease-enriched prompts refers to the concatenation of the name of the classes showing presence in one sample. Note that each sample may have multiple disease (class) labels, hence there might be dozens of disease-enriched prompt instances for each diseased-enriched prompt template by different presence of the disease information.

For instance, two samples with associated disease labels ``\textit{Lung Opacity, Edama}" and ``\textit{Lung Opacity, Atelectasis}" are provided with two disease-enriched prompt instances ``\textit{a picture of lung opacity, edama}" and ``\textit{a picture of lung opacity, atelectasis}" for $de3$ disease-enriched prompt template.

\begin{table}
\caption{The experimental results of ablation studies on the MIMIC-CXR datasets.}
\normalsize
\centering
\label{tab:prompt_results}
\begin{tabular}{c|c|cccccccc}

\toprule  
\textbf{ Knowledge } &\textbf{ Methods }  & \textbf{ BLEU-1} & \textbf{BLEU-2} & \textbf{BLEU-3} &
\textbf{BLEU-4} & \textbf{METEOR} & \textbf{CIDEr }  \\
\midrule  
\multirow{2}{*}{common} &+ cmn3 &31.9 &19.4 &13.2 &9.7 &13.1 &19.9   \\
\multirow{2}{*} &+ cmn4  &31.9  &19.3 &13.1 &9.5 &13.1 & 19.9  \\
\hline
\multirowcell{2}{domain\\ related} &+ dr3  &31.3 &18.8 &12.7 &9.2 &12.8 & 17.3 \\
\multirow{2}{*} &+ dr4  &31.1 &18.7 &12.5 &9.1 &12.8 & 18.2 \\
\hline
\multirowcell{4}{disease\\ enriched} &+ de3  &34.1 &21.4 &14.8 &11.2 &14.1 &25.9  \\
\multirow{4}{*} &+ de4  &34.0 &21.4 &14.9 &11.1 &14.2 &26.3  \\
\multirow{4}{*} &+ de5  &33.5 &20.9 &14.5 &10.7 &13.9 &25.0  \\
\multirow{4}{*} &+ de6 &34.0 &21.1 &14.6 &10.8 &14.0 & 24.4   \\
\multirow{4}{*} &+ de7 &34.1 &21.4 &14.8 &11.0 &14.2 &26.4   \\


\bottomrule 
\end{tabular}
\end{table}

\begin{table}
\small
\caption{Disease-enriched manual prompts generated by the LLM.}
\begin{center}
\aboverulesep=0ex 
\belowrulesep=0ex 
\begin{tabular}{c|c|c}
\toprule
\textbf{ Information } &\textbf{ Label } & \textbf{Manual Prompt $\bm{p}$}\\
\midrule
Instruction &-  & \begin{tabular}[c]{@{}l@{}}
You are a prompt generator. Firstly, I will give you a subject:\\ “disease enriched prompts prepending radiology reports”,\\ then you will generate 10 prompts such as: “a radiology\\ image presenting [disease name] that shows” or “a picture \\presenting  {[disease name]} that shows”. My first subject is  \\“disease enriched prompts prepending radiology reports”
\end{tabular}  \\
\hline
\multirowcell{4}{disease\\ enriched \\LLM} &de1-llm & \textit{a radiology report revealing signs of [class name] in the}  \\
\multirow{4}{*} &de2-llm & \textit{an x-ray image demonstrating [class name] often found in}  \\
\multirow{4}{*} &de3-llm & \textit{radiographic findings indicative of [class name] can be seen in this}  \\
\multirow{4}{*} &de4-llm & \textit{ a comprehensive x-ray study on the condition of [class name] depicting} \\
\multirow{4}{*} &de5-llm & \textit{ an in-depth radiology analysis of the [class name] syndrome, exhibiting} \\
\bottomrule
\end{tabular}
\end{center}
\label{tab:manual_prompt_llm}
\vspace{-30pt}
\end{table}

\section{The Performance of levering LLM as Prompt Engineer}

We also explore the performance of leveraging LLM as a prompt engineer to form the disease-enriched prompts \cite{zhou2022large}. Specifically, we give the large language model (LLM) GPT3.5~\cite{brown2020language} the following instruction as shown in the first row in \autoref{tab:manual_prompt_llm}. Then, we select five most related disease-enriched prompts generated by the GPT3.5. 

As shown in \autoref{tab:manual_prompt_llm}, the disease-enriched prompt generated by the LLM achieves comparable results to some manually designed prompts, but are still inferior to some best-performed manual prompts such as \textbf{de1} and \textbf{de2}, as presented in Table 3 of the main text. Note that we still need to give the LLM our defined disease-enriched prompt as the context to help the LLM model comprehend the task. 

Note that this work aims to utilize prompt learning and pre-trained language models to improve the RRG quality. Exploring the performance with general prompt learning methods is not the goal and therefore is currently out of the scope of the work.
However, we agree that more sophisticated approaches may yield even better performance gains and it would be worth investigating how best to leverage the LLM as a prompt engineer.

\begin{table}
\caption{The experimental results of the disease-enriched manual prompts generated by the GPT3.5.}
\normalsize
\centering
\label{tab:prompt_results}
\begin{tabular}{c|c|cccccccc}

\toprule  
\textbf{ Knowledge } &\textbf{ Methods }  & \textbf{ BLEU-1} & \textbf{BLEU-2} & \textbf{BLEU-3} &
\textbf{BLEU-4} & \textbf{METEOR} & \textbf{CIDEr }  \\
\toprule
\multirowcell{4}{disease\\ enriched \\ LLM} &+ de1-llm  &34.1 &21.4 &14.8 &11.0 &14.2 &26.3  \\
\multirow{4}{*} &+ de2-llm  &33.9 &21.3 &14.8 &11.0 &13.9 &27.7  \\
\multirow{4}{*} &+ de3-llm &33.5 &21.0 &14.6 &10.8 &14.0 & 26.5   \\
\multirow{4}{*} &+ de4-llm &33.9 &21.2 &14.7 &10.8 &14.2 &25.2   \\
\multirow{4}{*} &+ de5-llm &33.9 &21.2 &14.7 &10.9 &14.1 & 26.5   \\


\bottomrule 
\end{tabular}
\end{table}

\section{The Performance of More LLMs}

To further explore the influence of the large language models pretrained on medical domain and natural scene domain, we conduct further experiments on different large language models with the disease-enriched manual prompt ``de1". The experimental results are presented in \autoref{tab:more_llms}. 

In detail, we first explore the performance of our proposed methods on two medical-specific language models, i.e., ClinialBERT \cite{alsentzer2019publicly} and CXRBERT \cite{boecking2022making}. ClinialBERT is a medical-specific BERT trained on MIMIC-III v1.4 database \cite{johnson2016mimic} containing approximately 2 million clinical notes. CXRBERT further adds a chest radiology dataset MIMIC-CXR \cite{johnson2019mimic} in training to better understand the chest domain knowledge. Furthermore, two widely used natural scene domain text encoders \cite{he2020deberta,radford2019language} are employed to further investigate the sensitivity of our methods on different text encoders. 

From \autoref{tab:more_llms}, it can be seen that CXRBERT achieves slightly better results than ClinicalBERT on 3 out of 6 evaluation metrics and the same figures on the remaining 2 out of 3 metrics due to the better adaption to the chest radiology domain. 

Nonetheless, PromptRRG with text encoders trained in the traditional scene domain are shown to outperform  medical-specific encoders when generating reports. From the above results and the results in Table 3 in the main text, we can conclude that leveraging pretrained language models with prompt learning, simply increasing domain knowledge cannot demonstrate notably superior performance than the common prompts. We have shown that it is of great importance to strike the balance between the better encoding capability normally shown in a natural scene text encoder with common prompts and providing more medical information for the clinical text encoder or domain-related prompts.

More importantly, all the five LLMs demonstrate remarkable improvements when used with our proposed framework, confirming the effectiveness of the PromptRRG. Furthermore, the above results also show that our method is insensitive to using different LLMs.

\begin{table*}
\caption{The performance of different LLMs as text encoder. The best values are highlighted in bold. 
}
\centering
\label{tab:more_llms}
\begin{tabular}{c|c|cccccc}
\toprule  
\textbf{Domain} &\textbf{Text Encoder} & \textbf{BLEU-1} & \textbf{BLEU-2} & \textbf{BLEU-3} &
\textbf{BLEU-4}  & \textbf{METEOR} &\textbf{CIDEr}\\
\midrule
 \multirow{4}{*}{Medical} &$ClinialBERT$ \cite{vinyals2015show} &31.1 &18.9 &12.8 &9.3  &12.7 &22.6 \\
 \multirow{4}{*} &$ClinialBERT$ + de1  &34.1 &21.2 &14.6 &10.8  &13.9 & 27.0 \\
 \cline{2-8}
  \multirow{4}{*} &$CXRBERT$ \cite{rennie2017self} &31.9 &19.4 &13.2 &9.6  & 13.0 &23.8\\
  \multirow{4}{*} &$CXRBERT$ + de1 &34.1 &21.4 &14.8 &10.8  & 14.2 &22.9\\
 \toprule  
\multirowcell{6}{Natural \\ Scene} &$Deberta$ \cite{he2020deberta} &31.1 &18.8 &12.7 &9.3  &12.8 &23.4\\
\multirow{6}{*} &$Deberta$ + de1 &34.3 &21.6 &15.1 &11.2  & 14.3 &27.8\\
 \cline{2-8}
\multirow{6}{*} & $GPT2$ ~\cite{radford2019language} &30.3 &18.1 &12.2 &8.8  & 12.5 &19.1 \\
\multirow{6}{*} &$GPT2$ + de1 &34.4 &21.7 &15.1 &11.2  & 14.3 &26.8 \\
 \cline{2-8}
\multirow{6}{*} &$Roberta$ ~\cite{liu2019roberta}) &31.7 &19.3 &13.0 &9.4 &13.0 & 19.5 \\
\multirow{6}{*} &$Roberta$ + de1 &\textbf{34.8} &\textbf{21.9} &\textbf{15.3} &\textbf{11.3} & \textbf{14.5} &\textbf{28.6}\\

\bottomrule 
\end{tabular}
\vspace{-10pt}
\end{table*}

\section{The Performance of a Visual-Language Model}

We also tested our method with a recent state-of-the-art visual language model, i.e., METER~\cite{dou2022empirical}. As shown in \autoref{tab:vl_results}, our proposed method also notably improves the base METER model by a large margin. Nonetheless, a pretraiend visual-language model cannot bring any improvements compared with the decoder-free (decoder trained from scratch) method as shown in the last row in \autoref{tab:vl_results}. We conjecture that this is because the hidden size in the decoder in the pretrained visual-language model, i.e., 768 with 12 attention heads, is too large and may cause the overfitting problem on RRG. The decoder of the proposed PromptRRG consists of three vanilla transformer blocks with a hidden size of 512 and eight attention heads. We further conducted an experiment on 768 hidden size with 12 attention heads settings (\textbf{Our-768)}. As shown in the last two rows in \autoref{tab:vl_results}, a significant performance drop can be seen when increasing the hidden size to 768, possibly confirming our conjecture.

\begin{table*}
\caption{The performance of a visual-langugae model METER. The best values are highlighted in bold. 
}
\centering
\label{tab:vl_results}
\begin{tabular}{cccccccc}
\toprule  
\textbf{Method} & \textbf{BLEU-1} & \textbf{BLEU-2} & \textbf{BLEU-3} &
\textbf{BLEU-4}  & \textbf{METEOR} &\textbf{CIDEr}\\
\midrule

  $METER$ \cite{dou2022empirical} &30.8 &19.0 &12.5 &8.8  &12.2 & 9.6 \\
 $METER$ + de1 &33.8 &21.2 &14.7 &11.0  &26.0 & 14.1 \\

\midrule  
\textbf{Our-768} (decoder-free) &{33.9} &{21.1} &{14.6} &{10.7} & {23.7} &{14.1}\\
\textbf{Our} (decoder-free) &\textbf{34.8} &\textbf{21.9} &\textbf{15.3} &\textbf{11.3} & \textbf{14.5} &\textbf{28.6}\\

\bottomrule 
\end{tabular}
\vspace{-10pt}
\end{table*}


\clearpage

{\small
\bibliographystyle{ieee}
\bibliography{supple/supplebib}
}